\documentclass[letterpaper]{article} 
\usepackage{aaai24}  
\usepackage{times}  
\usepackage{helvet}  
\usepackage{courier}  
\usepackage[hyphens]{url}  
\usepackage{graphicx} 
\usepackage{natbib}  
\usepackage{caption} 
\usepackage{algorithm}
\usepackage{algorithmic}

\usepackage{amsmath}
\usepackage{amssymb}
\usepackage{multirow}

\title{TEILP: Time Prediction over Knowledge Graphs via Logical Reasoning}
\author{
    Siheng Xiong\textsuperscript{\rm 1},
    Yuan Yang\textsuperscript{\rm 1},
    Ali Payani\textsuperscript{\rm 2},
    James C Kerce\textsuperscript{\rm 1},
    Faramarz Fekri\textsuperscript{\rm 1} 
}
\affiliations{
  \textsuperscript{\rm 1}  Georgia Institute of Technology, 225 North Avenue, Atlanta, Georgia 30332 USA\\
    \textsuperscript{\rm 2} Cisco Systems Inc, 300 East Tasman Dr, San Jose, California 95134 USA\\
    \{sxiong45, yyang754\}@gatech.edu, apayani@cisco.com, \\ clayton.kerce@gtri.gatech.edu,  faramarz.fekri@ece.gatech.edu

}

\begin{document}

\maketitle

\begin{abstract}
Conventional embedding-based models approach event time prediction in temporal knowledge graphs (TKGs) as a ranking problem. However, they often fall short in capturing essential temporal relationships such as order and distance. In this paper, we propose TEILP, a logical reasoning framework that naturally integrates such temporal elements into knowledge graph predictions. We first convert TKGs into a temporal event knowledge graph (TEKG) which has a more explicit representation of time in term of nodes of the graph. The TEKG equips us to develop a differentiable random walk approach to time prediction. Finally, we introduce conditional probability density functions, associated with the logical rules involving the query interval, using which we arrive at the time prediction. We compare TEILP with state-of-the-art methods on five benchmark datasets. We show that our model achieves a significant improvement over baselines while providing interpretable explanations. In particular, we consider several scenarios where training samples are limited, event types are imbalanced, and forecasting the time of future events based on only past events is desired. In all these cases, TEILP outperforms state-of-the-art methods in terms of robustness.
\end{abstract}

\section{Introduction}

Temporal knowledge graphs (TKGs) are an important representation when dealing with dynamic and time-dependent relationship between entities. They have various applications such as healthcare and medical research, social event analysis, and recommendation systems. TKGs contain the quadruple $(e_s, P, e_o, t)$ describing the relation $P$ between subject entity $e_s$ and object entity $e_o$ at time $t$.  Due to their large scales, real-world TKGs usually suffer from incompleteness. Thus, link prediction and time prediction, i.e., inferring missing entity and time with existing facts, are the common reasoning tasks for TKGs, proposed in either formal structured query or natural language \cite{saxena2021question} \cite{chen-etal-2023-multi}. Compared with link prediction, time prediction is even more challenging as a regression task \cite{cai2022temporal}.

Existing embedding-based methods consider time prediction as a ranking problem, e.g., HyTE \cite{dasgupta2018hyte}, Time-Aware Embedding \cite{garcia-duran-etal-2018-learning}, DE-SimplE \cite{goel2020diachronic} and TNT-ComplEx \cite{Lacroix2020Tensor}. The underlying principle is to calculate the score of each timestamp given the triple of subject, object and relation. To predict intervals, TimePlex \cite{jain2020temporal} introduces a greedy coalescing strategy, which greedily extends the timestamp with the highest score into an interval. Although these methods give time predictions, they can neither handle unseen timestamps nor utilize the intrinsic connections between timestamps such as temporal order and distance.   

Alternatively, inductive logical reasoning methods, e.g., StreamLearner \cite{omran2019learning}, TLogic \cite{liu2022tlogic}, and TILP \cite{xiong2022tilp}, have desirable features when applied to TKGs, as they provide interpretable and robust inference results, and can easily incorporate external background knowledge and domain-specific rules into the reasoning. However, these qualitative predicates based logical rules alone are not enough for time prediction. Thus, several methods built on the temporal point process, e.g., Know-Evolve \cite{trivedi2017know}, GHNN \cite{han2020graph}, TLPP \cite{pmlr-v119-li20p} and TELLER \cite{li2021explaining}, have been introduced to solve the forecasting task, i.e., time prediction from previous events. Compared with the general setting, this task is restricted since all history events are required, and its target is to predict the time gap between the last known event and the next one. 

In this paper, we propose TEILP which converts TKGs into a \text{temporal event knowledge graph} (TEKG) and enables a \text{differentiable} random walk approach to solve the time prediction problem. In TEILP, for each learned rule, a conditional probability density function is associated with the query interval. This density function is then used in a Gaussian mixture model to predict the time. We achieve better performance than embedding methods while providing an additional benefit of human-readable logical explanations. More specifically, our main contributions are:

\begin{itemize}
    \item We propose TEILP, a temporal logical reasoning framework for time prediction. It is the \textbf{first} inductive approach that directly learns temporal logical rules and associated conditional probability density functions for the time prediction. 
    \item We introduce a novel \textbf{differentiable} temporal random walk approach by converting TKGs into TEKG where multi-type nodes denote either events or timestamps, and multi-type edges denote either entities or temporal relations.
    \item We achieve better performance than state-of-the-art baselines on five benchmark TKG datasets. Our model also demonstrates its robustness in several scenarios where training samples are limited, event types are imbalanced, and forecasting the time of future events based on only past events is desired.
\end{itemize}

\section{Related Work}

Time prediction over knowledge graph has been a challenging task since the timestamps come from a continuous space with intrinsic dependencies such as order and distance. Embedding-based methods, e.g., HyTE \cite{dasgupta2018hyte}, Time-Aware Embedding \cite{garcia-duran-etal-2018-learning}, DE-SimplE \cite{goel2020diachronic} and TNT-ComplEx \cite{Lacroix2020Tensor}, focus more on link predication, and consider time prediction as a similar ranking problem for different timestamps. HyTE projects the relation and entities on all the temporal hyperplanes, and orders the timestamps according to their plausibility scores. TNT-ComplEx creates time-dependent embeddings for entities and relations, and examines as to how the score of facts changing with time. To handle the interval prediction task, the authors of TimePlex \cite{jain2020temporal} introduces a greedy coalescing strategy, and extends embedding-based models via additional temporal constraints (relation recurrence, ordering and time gap distribution). When predicting time, both embedding-based scores and these constraints are considered to rank the timestamps. Time2box \cite{cai2021time} considers timestamps as box filters, picking out correct answers for temporal queries. However, embedding-based timestamp ranking methods cannot be very accurate since they ignore the inter-dependencies (such as order of events and time distance between the events), and cannot handle unseen timestamps.

Symbolic methods for temporal knowledge graph reasoning use only qualitative predicates, e.g. StreamLearner \cite{omran2019learning}, TLogic \cite{liu2022tlogic}, ALRE-IR \cite{mei2022adaptive}, TILP \cite{xiong2022tilp}, and thus have no capability of time prediction. The authors of NeuSTIP \cite{singh2023neustip} solve this problem by introducing a Gaussian distribution for the query timestamp. However, depending on both path embedding and rule head embedding, their method is not applicable to the inductive setting. In addition, there are several inefficiency in their approach: (i) The distribution parameters are only related to the head predicate and the first predicate in the rule body, and hence many different rule patterns share the same parameters. (ii) Given a path connecting the subject and object entity, they only use one timestamp from the first body predicate, ignoring other useful information. (iii) They assume to know the temporal relation between the head predicate and the first body predicate, which is unknown in time prediction. 

Another body of works looked into a related (forecasting) task of how to predict the time of next future event given all previous events. These works tend to use the temporal point process (TPP) for modelling. Embedding-based methods include Know-Evolve \cite{trivedi2017know}, GHNN \cite{han2020graph} and EvoKG \cite{park2022evokg}. Symbolic methods include TLPP \cite{pmlr-v119-li20p} and TELLER \cite{li2021explaining}. Compared with the general setting of Temporal Knowledge Graph Completion (TKGC), this forecasting task is more restrictive since not all history events are available in real-world applications. Due to the essence of causal sequence modelling, auto-regressive strategy, i.e., generating outputs based on past outputs and inputs, will also fail without the temporal order prior of the queries.

\section{Our Framework}

\subsubsection{Problem Definition} Let $\mathcal{E}$ be the set of entities, $\mathcal{P}$ be the set of predicates (or relations), $\mathcal{T}$ be the set of timestamps, $\mathcal{I} \subset \mathcal{T} \times \mathcal{T} $ be the set of intervals. A TKG $\mathcal{G} \subset \mathcal{E} \times \mathcal{P} \times \mathcal{E} \times \mathcal{I}$ is composed of quadruples $\left(e_s, P, e_o, I\right)$ where $e_s, e_o \in \mathcal{E}$ denote subject and object entities, respectively, $P\in \mathcal{P}$ denotes predicate (or relation), and $I\in \mathcal{I}$ denotes the time. Let $\left(e_s, P, e_o, ?\right)$ be the query. Then, our objective is to predict the time $I$ based on the observed facts from the same TKG $\mathcal{G}$. Compared with link predication, time prediction is more challenging as a regression (or interval estimation) task.

\begin{figure}[t]
    \centering
\includegraphics[width=0.47\textwidth]{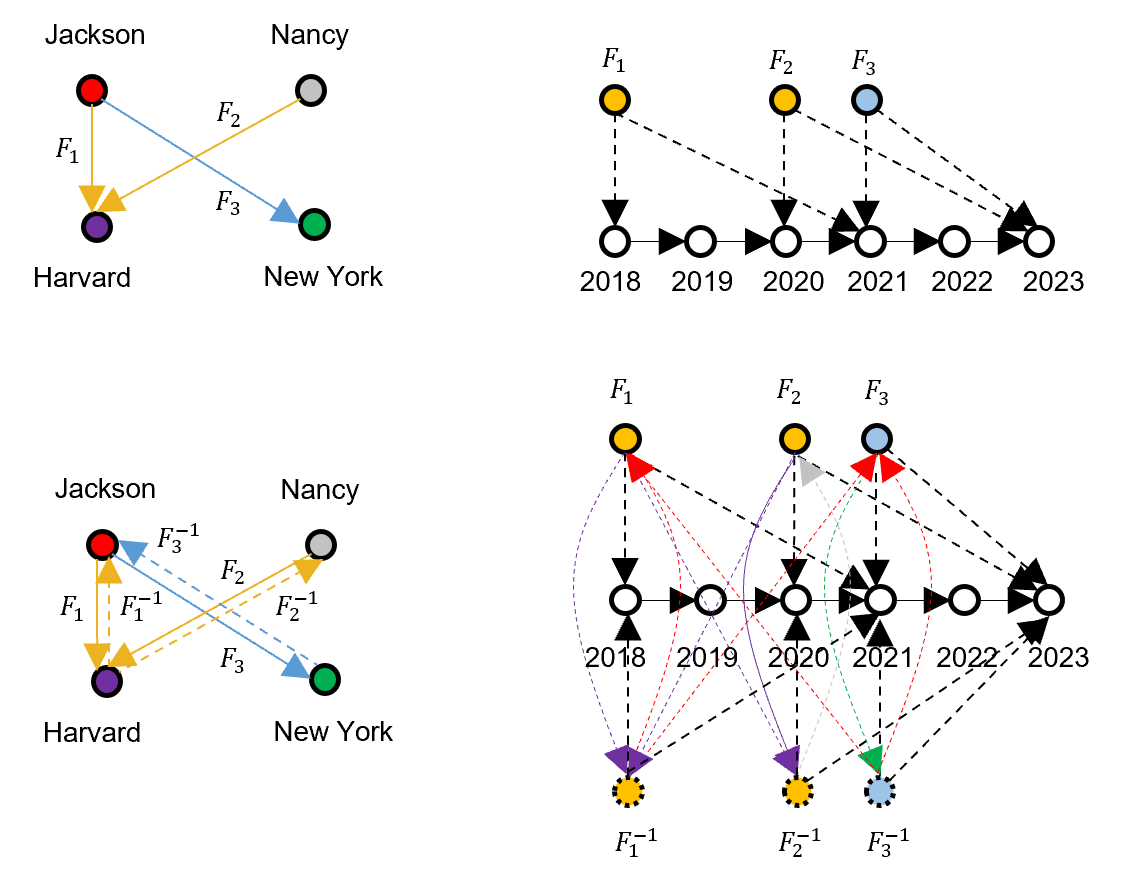}
  \caption{An example TKG (left) and the corresponding TEKG (right). The first row are the original versions, and the second row are the enhanced versions.}
  \label{fig:1}
\end{figure}

\begin{figure*}[t]
    \centering
    \includegraphics[width=0.8\textwidth]{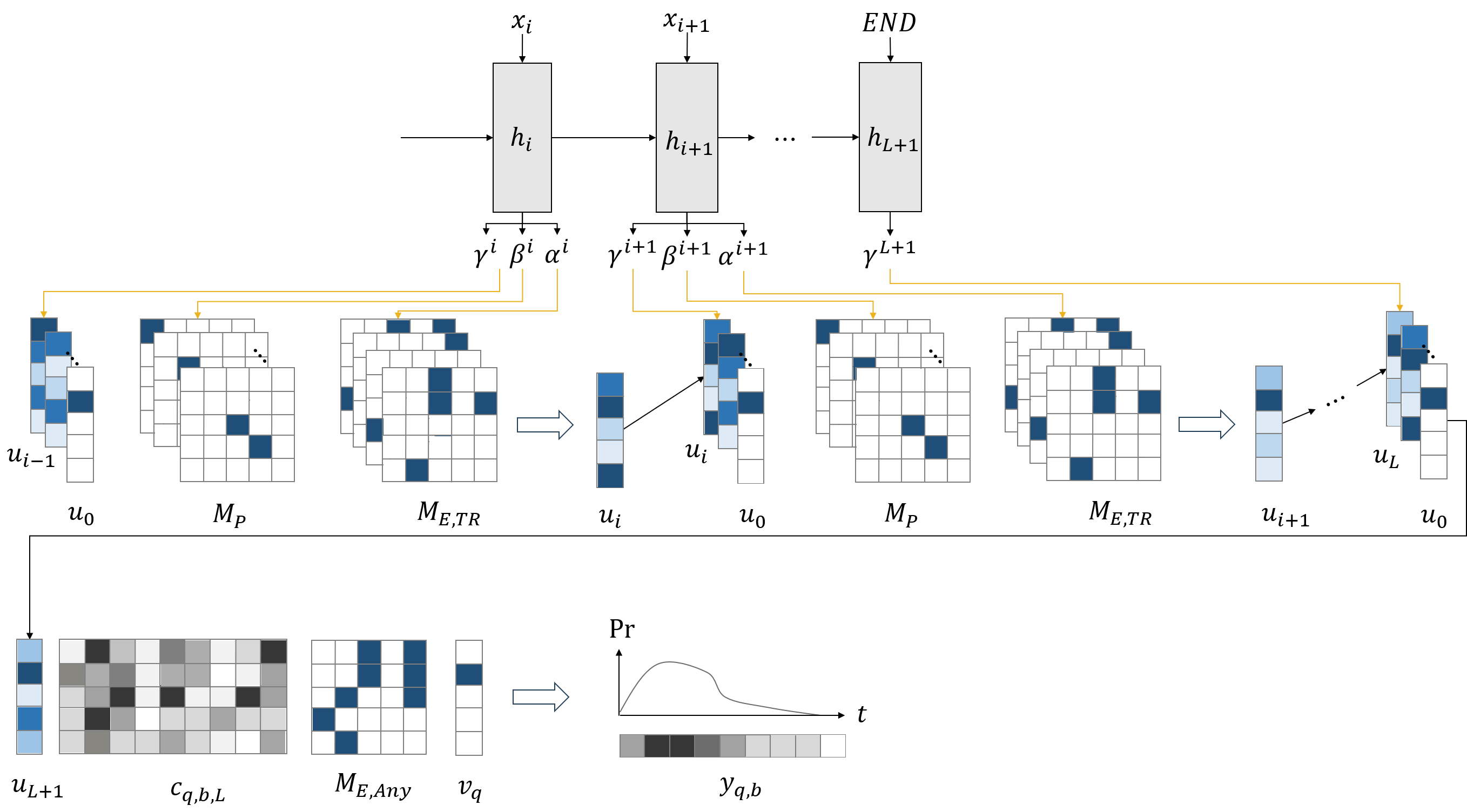}
  \caption{The illustration of rule-based time prediction.}
  \label{fig:2}
\end{figure*}

\subsubsection{Temporal Event Knowledge Graph} To solve the time prediction problem, given a TKG, we propose to convert the corresponding TEKG uniquely as the following. In TEKG, there exist two types of nodes: i) For each fact $(e_s, P, e_o, I)$ in TKG, we define a corresponding event node $F\in \mathcal{F}$, where $\mathcal{F}$ is the set of facts, i.e., $F := (e_s, P, e_o, I)$. ii) For each timestamp $t$ in TKG, a corresponding timestamp node $T$ is defined in TEKG. Given the node definition, we further define three types of edges: i) An entity edge $E_{FF^{\prime}}$ exists from $F$ to $F^{\prime}$ iff some entity $e\in \mathcal{E}$ is the object of $F$ as well as the subject of $F^{\prime}$; ii) A temporal order edge $E_{TT^{\prime}}$ exists between consecutive timestamp nodes $T$ and $T^{\prime}$; iii) A start time edge $E_{FT,s}$ or end time edge $E_{FT,e}$ exists from $F$ to $T$ iff $T$ is the start time or end time of $F$. Figure 1 visualizes an example TKG and the corresponding TEKG where $F_1 = (\text{'Jackson'},\text{'StudyIn'},\text{'Harvard'}, [2018,2021])$, $F_2 = (\text{'Nancy'},\text{'StudyIn'},\text{'Harvard'}, [2020,2023])$ and $F_3 = (\text{'Jackson'},\text{'WorkIn'},\text{'New York'}, [2021,2023])$.

Given the above definition, we describe an important property of TEKG: if there exists a path of entity edges between two event nodes, we can always find a corresponding path in TKG. This property ensures that we can find in TKG the equivalents of logical rules learned in TEKG. Further, we consider the common enhancement strategy of adding inverse edges in TKG. These inverse edges which interchange the position of subject and object entity are introduced to allow bi-directional random walks in TKG. Similarly, in TEKG, we define mirror nodes $F^{-1}$ to represent these inverse events. The entity edges and start time or end time edges of mirror nodes follow the same definition as original nodes. Figure 1 also visualizes the enhanced TKG and corresponding TEKG by adding inverse edges and mirror nodes, respectively. As the foundation of our approach, TEKG enables a differentiable random walk process. It allows us to better learn rule structure and confidence using gradient-based optimizer.

\subsubsection{Temporal Logical Rules in TEKG} A temporal logical rule of length $l \in \mathbb{N}$ in TEKG is defined as
\begin{equation}
\begin{split}
& Z_{R_P,I_1, \cdots, I_{l}}(I_q) \leftarrow E\left(F_q, F_1\right) \wedge \cdots \wedge E\left(F_{l-1}, F_{l}\right)  \\ &\quad \wedge E(F_{l}, F_{q}) \wedge P_q\left(F_q\right) \wedge P_1\left(F_1\right)\wedge \cdots \wedge P_{l}\left(F_{l}\right) \\
&\quad \wedge TR_1(I_1, I_2) \wedge \cdots \wedge TR_{l-1}(I_{l-1}, I_{l})
\end{split}
\end{equation}
where $F_q$ denotes the query event, and $I_q$ denotes its interval. $\{F_i\}_{i\in\mathbb{N}}$ denote the variables of fact, and $\{{I}_i\}_{i\in\mathbb{N}}$ denote their interval. We ground these variables during inference. To allow generalization, $E(\cdot)$ denotes an entity edge related to any entity $e\in\mathcal{E}$, and $P_i(\cdot)$ denotes a grounded predicate, $TR_{i}(\cdot)\in \{\text{Before}$, $\text{Overlap}$, $\text{After}$, $\text{Any}\}$ denotes a grounded temporal relation between two intervals (timestamps), which is defined in \cite{xiong2022tilp}. Predicate $Z(\cdot) \in \{0, 1\}$ acts as an indicator on its arguments in relation to the rule $R_P := [P_q, P_1, \cdots, P_{l}, TR_1, \cdots, TR_{l-1}]$ and relevant intervals $\{I_1, \cdots, I_l\}$. 

The left arrow in rule is called “implication", i.e., the rule body on the right implies the rule head on the left. The rule head $Z_{R_P, I_1, \cdots, I_{l}}(\cdot)$ indicates whether $I_q$ satisfy $R_P$ given the relevant intervals $\{I_1, \cdots, I_l\}$. The rule body contains the query predicate $P_q(\cdot)$, which is given in time prediction, and a cyclic path involving $F_q$, specified by predicate $P_i(\cdot)$ and temporal relation $TR_i(\cdot)$. The intuition here is to use logical rules to help us find $l$ relevant events $\{F_1, \cdots, F_{l}\}$ for time prediction of the query $F_q$.

\subsection{Logical Reasoning via Random Walk} Given a logical rule, the grounding process is to replace variables into constant terms. If the structure of rule body can be corresponded to a path in knowledge graph, the inference is equivalent to performing random walk under some constraints. In our temporal logical rules, there are three types of constraints: connectivity, predicates and temporal relations. We first design the operator for node attributes. Given the predicate $P(\cdot)$, for every node $F_{m} \in \mathcal{F}$, the operator $M_{P}\in\{0,1\}^{|\mathcal{F}| \times|\mathcal{F}|}$ is defined such that its $(m, m)$ entry is 1 iff $P(F_m)$ is true. We then define the operator for edge attributes. Given the connectivity $E(\cdot)$ and temporal relation $TR(\cdot)$, for every pair of events $F_m, F_n \in \mathcal{F}$, the operator $M_{E,TR}\in\{0,1\}^{|\mathcal{F}| \times|\mathcal{F}|}$ is defined such that its $(n, m)$ entry is 1 iff $\exists e \in \mathcal{E}$, s.t. both $E(F_m, F_n)$ and $TR(I_m, I_n)$ are true, where $I_m$ and $I_n$ are the time of $F_m$ and $F_n$, respectively. Note that, we use a single operator for connectivity $E$ and temporal relation $TR$, which implicitly implies a logical ``and" in the rules. Compared with using separate operators, this strategy is more efficient in practice.

Given the operators, we define a \textbf{differentiable} temporal random walk with a recurrence formulation:
\begin{equation}
    \mathbf{v}_{i+1} = \sum_{j=1}^{|TR|} \alpha_j^i M_{E,TR_{j}} \left(\sum_{k=1}^{|P|} \beta_k^i M_{P_k} \mathbf{v}_{i}\right)
\end{equation}
where $\mathbf{v}_i\in[0,1]^{{|\mathcal{F}|}}$ is the state vector for step $i$, representing the probability distribution of different events. $M_{P_k}$ with the predicate index $k$ is the operator for $P_k$, and $\beta_k^i$ is the weight for $P_k$ at step $i$. Here we use a soft selection from different predicates: $\forall i$, $\beta_k^i\in[0,1]$ and $\sum_{k=1}^{|P|} \beta_k^i=1$. Similarly, $M_{E, TR_j}$ with the temporal relation index $j$ is the operator for both $E$ and $TR_j$, and $\alpha_j^i$ is the weight for $TR_j$ at step $i$. We have $\forall i$, $\alpha_j^i\in[0,1]$ and $\sum_{j=1}^{|TR|} \alpha_j^i=1$ as a soft selection from different temporal relations.

\subsection{Time Prediction}
To use inductive rules in (1) for time prediction, we introduce a conditional probability density function $G(\cdot)$ which describes the relationship between $I_q$ and $\{I_1, \cdots, I_l\}$. There exist multiple choices for $G(\cdot)$. In this paper, we consider modelling the time gap between the query timestamp in $I_q$ and the known timestamp in $\{I_1, \cdots, I_l\}$. The intuition is that the time gap shares the same probability distribution among different events. For example, the time gap between the same person's birth date and death date, i.e., a person's lifespan, follows a Gaussian distribution, across different persons. Similarly, the time gap between the same person's birth date and university graduation date follows another Gaussian distribution. Further, we consider these distributions evolving with time, e.g., the lifespan of modern people is significantly longer than that of ancient humans.

To be specific, the relationship between $I_q := [t_{q,s}, t_{q,e}]$ and $\{I_1 := [t_{1,s}, t_{1,e}]$, $\cdots$, $I_l := [t_{l,s}, t_{l,e}]\}$ is defined as:
\begin{equation}
    G_{R_P, b}(t_{q,b}|I_1, \cdots, I_l) = \sum^l_{i=1} \ a_{P_q, b, i}\cdot g_{R_P,b,i}(t_{q,b}|I_i) 
\end{equation}
where $G(\cdot)$ denotes the conditional probability density related to a pattern $R_P$ and subscript $b = \{\text{'s'}, \text{'e'}\}$. Note that, (3) updates $G$ whenever the rule in (1) is satisfied by the temporal pattern of events, i.e., $R_P: Z_{R_P,I_1, \cdots, I_{l}}(I_q) = 1$. The components $g(\cdot)$ denote the conditional probability density related to $R_P$, subscript $b$ and index $i$ with learnable weights $a_{P_q, b, i}\in[0,1]$ and $\sum^l_{i=1} \ a_{P_q, b, i}=1$, where $P_q$ denotes the query predicate.

\begin{equation}
\begin{aligned}
    g_{R_P,b,i}(t_{q,b}|I_i) & = \ w_{P_q, b, i} \cdot f_{R_P,b,i,s}\left(t_{q,b}-t_{i,s}\right) + \\ & \left(1-w_{P_q, b, i}\right)\cdot f_{R_P,b,i,e}\left(t_{q,b}-t_{i,e}\right)
\end{aligned}    
\end{equation}

\begin{algorithm} [t]
\caption{: Rule Learning}
\label{alg:rule_learning}
\begin{algorithmic}[1]
\item[]\textbf{Input:} Temporal knowledge graph $\mathcal{G}$, query event $F_q$, target interval $I_q := [{t}_{q,s}, {t}_{q,e}]$.
\item[] \textbf{Parameters:} Maximum rule length $L$, flag for duration modelling $\mathbb{F}_{t_d}$.
\item[] \textbf{Output:} Rule patterns $S_{R_P}$, probability density functions $g_{R_P, b, i}(\cdot)$, attention vectors $\alpha, \beta, \gamma$.
\STATE Convert $\mathcal{G}$ into temporal event knowledge graph $\mathcal{G}^{\prime}$.
\STATE Sample cyclic random walks of length $\l \in [1, L]$ on $\mathcal{G}^{\prime}$ starting from either $F_q$ or $F_{q^{-1}}$ to obtain rule pattern candidates $S_{R_P}$ and local graph $\mathcal{G}^{\prime}_q$.
\STATE For each rule pattern $R_P \in S_{R_P}$, fit the probability density functions $g_{R_P, b, i}(\cdot)$ for subscript $b \in \{$\text{'s'}, \text{'e'}, \text{'d'}\} and index $i \in \{1, l\}$. 
\STATE For all known events $F_m \in \mathcal{G}^{\prime}_q$ and target timestamp $t_{q,b}$, calculate $g_{R_P, b, i}(t_{q,b} | I_m)$ if $F_m$ satisfy $R_P$.
\STATE Calculate the probabilities $\operatorname{Pr}(t_{q,b})$ based on either (5) - (9) (event-split version) or (10) (rule-split version).
\STATE Learn the optimal attention vectors $\alpha, \beta, \gamma$ from (11).
\end{algorithmic}
\end{algorithm}

\noindent where the components $f(\cdot)$ denotes the conditional probability density related to $R_P$, $b$, $i$ and subscript 's' or 'e' with learnable weights $w_{P_q, b,i}\in[0,1]$. More details for the probability density function design are shown in the supplementary material. During training, functions $f(\cdot)$ will be fitted, and weights $a$ and $w$ will be learned. Further, we consider two options for modelling ${t}_{q,e}$: estimate it directly, or estimate duration ${t}_{q,d}$ and set $\widehat{t}_{q,e} := \widehat{t}_{q,s} + \widehat{t}_{q,d}$. We compare them in experiments.

\subsection{Rule Learning \& Application}
The pseudocode for rule learning is described in Algorithm 1. Inspired by Neural-LP \cite{yang2017differentiable}, we use an attention mechanism to deal with varying rule length. Compared with Nerual-LP, which is developed for static link prediction, we add operators for connectivity, temporal relation and conditional probability density functions for time prediction.
\begin{align}
& \mathbf{u}_{1} = M_{E,\text{Any}} \mathbf{u}_{0}\\
& \mathbf{u}_{i} = \sum_{j=1}^{|TR|} \alpha_j^i M_{E,TR_{j}} \left(\sum_{k=1}^{|P|} \beta_k^i M_{P_k} \left(\sum_{\tau=0}^{i-1} \gamma^i_{\tau} \mathbf{u}_\tau\right)\right)\\
& \mathbf{u}_{L+1} = \sum_{\tau=0}^L \gamma^{L+1}_{\tau} \mathbf{u}_\tau\\
& \mathbf{y}_{q,b} = \mathbf{v}^{\text{T}}_q M_{E,\text{Any}} \left(\mathbf{c}_{q,b,L} \odot h_{\text{RPT}}\left(\mathbf{u}_{L+1}, |S_t| \right)\right)
\end{align}
where $\mathbf{u}_i \in [0,1]^{|\mathcal{F}|}$ denotes the  partial inference result at step $i$. The recurrent formulation of (6) is based on (2). The difference is that we use an attention vector, i.e., $\forall i, \gamma^i_{\tau}\in [0,1]$ and $\sum_{\tau=0}^{i-1} \gamma_{\tau}^i=1$, to softly select previous inference results. The initial result in (5) is the one-hot encoding of the query event, i.e., $\mathbf{u}_0 = \mathbf{v}_q\in\{0,1\}^{|\mathcal{F}|}$ whose q-th entry is 1 only. Further, $M_{E,\text{Any}}$ is the matrix operator for connectivity $E$ and temporal relation 'Any'. Finally, $L\in \mathbb{N}$ denotes the maximum rule length. The inference result $\mathbf{u}_{L+1}$ in (7) is a soft selection from all the previous results $\{\mathbf{u}_{0}, \mathbf{u}_{1}, \cdots, \mathbf{u}_{L}\}$. It represents the probability distribution we arrive at different events after at most $L$-step random walk. We calculate the time prediction $\mathbf{y}_{q,b}\in[0,1]^{|S_t|}$, where subscript $b\in\{\text{'s'}, \text{'e'}, \text{'b'}\}$, with (8). To allow efficient matrix operations, we quantize the timestamp range $[t_{min}, t_{max}]$ into a set of timestamps $S_t:=\{t_{min}, \cdots, t_r, \cdots, t_{max}\}$. In experiments, we use a uniform discretization, and more complex quantizations can be adopted. Conditional probability matrix $\mathbf{c}_{q,b,L}\in[0,1]^{|\mathcal{F}|\times|S_t|}$ is based on the conditional probability density function $g_{R_P, b, l}(\cdot)$. The detailed calculation of $\mathbf{c}_{q,b,L}$ is given in the supplementary material. Function $h_{\text{RPT}}(\cdot)$ duplicates $\mathbf{u}_{L+1}$ along axis 1 such that $h_{\text{RPT}}\left(\mathbf{u}_{L+1}, |S_t| \right)\in[0,1]^{|\mathcal{F}|\times|S_t|}$. Operator $\odot$ denotes an element-wise multiplication, and the left part $\mathbf{v}^\text{T}_qM_{E,\text{Any}}$ is introduced since we require the path to return to $F_q$. Note that, we only use the last event on the path for prediction. In experiments, we found that middle intervals $\{I_2, \cdots, I_{l-1}\}$ have a less significant impact on the performance. To involve the first event on the path, a trick here is to replace the query event $F_q$ with its mirror node $F_{q^{-1}}$. Let $\mathbf{y}_{q,b}$ and $\mathbf{y}_{q^{-1},b}$ be the corresponding time predictions. Based on (3), the final time prediction result can be written as:    
\begin{equation}
    Y_{q,b} = a_{P_q,b}\cdot\mathbf{y}_{q,b} + (1-a_{P_q,b})\cdot\mathbf{y}_{q^{-1},b}
\end{equation}
where the learnable weight $a\in[0,1]$ is related to the query predicate $P_q$ and subscript $b$.

Further, (5) - (9) essentially provide an \textbf{event-split} version for time prediction. Given the rules, we first calculate the probability of arriving at different events, and then predicts the query given the interval of the events. Alternatively, the \textbf{rule-split} version is to directly predict the query given the rule confidence and the interval of the events satisfying the rule, i.e.,
\begin{equation}
\small
\begin{aligned}
& \left(Y_{q,b}\right)_r = \sum_{\kappa=1}^{\left|S_{R_p}\right|} \textbf{s}_{R_p^\kappa}\left(\alpha, \beta, \gamma\right)(\left|S_{p ath}^\kappa\right|)^{-1}\sum_{\zeta_\kappa =1}^{\left|S_{path}^\kappa\right|} \left(a_{P_q, b} \cdot  \right.\\
&\quad \left.g_{R^\kappa_P,b,1}(t_r \mid I^{\zeta_\kappa}_1) + (1- a_{P_q, b}) \cdot g_{R^\kappa_P,b,l_\kappa}(t_r \mid I^{\zeta_\kappa}_{l_\kappa}))\right)
\end{aligned} 
\end{equation}

\noindent where $Y_{q,b}\in[0,1]^{|S_t|}$ denotes the time prediction, and $(Y_{q,b})_r$ denotes its $r$-th entry corresponding to candidate $t_r \in S_t$. Rule $R_P$ is indexed by $\kappa$, and $S_{R_p}$ is the set of rules. The rule score function $\mathbf{s}(\cdot)$ is defined in \cite{yang2017differentiable}. Further, $S_{path}^\kappa$ is the set of paths given $F_q$ and $R^\kappa_P$. Learnable weight $a\in [0,1]$ is conditioned on query predicate $P_q$ and subscript $b$. Finally, $I^{\zeta_\kappa}_1$, $I^{\zeta_\kappa}_{l_\kappa}$ are the corresponding intervals given the $\zeta_\kappa$-th path with length $l_\kappa$. 

Based on previous analysis, we know that the task of learning temporal logical rules is to learn the attention vectors $\mathbf{\alpha, \beta, \gamma}$ which softly select predicates, temporal relations and rule lengths, respectively. Inspired by TILP \cite{xiong2022tilp}, we use an LSTM model illustrated in Figure 2 to ensure that current step's attention vectors depend on previous steps'. The calculation is given in the supplementary material. Further, to ensure the efficiency of our model on large TKGs, we adopt some acceleration strategies and analyze the time complexity in the supplementary material. 

Training of the model is to minimize the log-likelihood loss:

\begin{algorithm} [t]
\caption{: Rule Application}
\label{alg:rule_application}
\begin{algorithmic}[1]
\item[]\textbf{Input:} Temporal knowledge graph $\mathcal{G}$, query event $F_q$.
\item[] \textbf{Parameters:} Rule patterns $S_{R_P}$, probability density functions $g_{R_P, b, i}(\cdot)$, attention vectors $\alpha, \beta, \gamma$, flag for duration modelling $\mathbb{F}_{t_d}$, quantized time range $S_t$.
\item[] \textbf{Output:} Predicted interval $\widehat{I}_q$.
\STATE Convert $\mathcal{G}$ into temporal event knowledge graph $\mathcal{G}^{\prime}$.

\STATE Given $S_{R_P}$, obtain a local knowledge graph $\mathcal{G}^{\prime}_q$ via cyclic walks starting from either $F_q$ or $F_{q^{-1}}$ on $\mathcal{G}^{\prime}$.

\STATE For all known events $F_m \in \mathcal{G}^{\prime}_q$ and candidate timestamps $t_r\in S_t$, calculate $g_{R_P, b, i}(t_r | I_m)$ if $F_m$ satisfy $R_P$.

\STATE Calculate predictions $Y_{q,b}$ with $\alpha, \beta, \gamma$ based on either (5) - (9) (event-split version) or (10) (rule-split version).

\STATE Estimate $\widehat{t}_{q,s}$ with $Y_{q,s}$ based on (12).

\STATE {If} $\mathbb{F}_{td}$ = True, estimate $\widehat{t}_{q,d}$ with $Y_{q,d}$ and set $\widehat{t}_{q,e} = \widehat{t}_{q,s} + \widehat{t}_{q,d}$, {else} directly estimate $\widehat{t}_{q,e}$ with $Y_{q,e}$.

\STATE Set $\widehat{I}_q = [\widehat{t}_{q,s}, \widehat{t}_{q,e}]$.
\end{algorithmic}
\end{algorithm}

\begin{equation}
\small
    \mathcal{L} = -\sum_{F_q}  \left(\log \operatorname{Pr}\left(t_{q,s} \mid Y_{q,s}\right) +\log \operatorname{Pr}\left(t_{q,e} \mid Y_{q,e}\right)\right)
\end{equation}
where $\operatorname{Pr}(t_{q,b}\ |Y_{q,b})$ denotes the probability of $t_{q,b}$ given the prediction $Y_{q,b}\in[0,1]^{|S_t|}$.

The pseudocode for rule application is described in Algorithm 2. Given the learned rule patterns $S_{R_P}$, probability density functions $g_{R_P, b, i}(\cdot)$ and attention vectors $\alpha, \beta, \gamma$, inference of the model is to find the timestamp $t_r$ in $S_t$ that maximizes the probability:

\begin{equation}
\widehat{t}_{q,b} = \underset{t_r \in S_t}{\operatorname{argmax}} \operatorname{Pr}\left(t_r \mid Y_{q,b}\right)
\end{equation}

\noindent The underlying logic of our method is to model probability distribution of the query interval. An alternative strategy is to directly perform regression. We found in experiments that the regression-based approach is essentially memorizing the answer. Their performance becomes much worse in the future event time forecasting.

\section{Experiments}

\begin{table*}[h]
\label{table-1}
\begin{center}
\renewcommand{\arraystretch}{1.05}
\begin{tabular}{l|cc|cc|c|c|c}
\hline
\multicolumn{1}{l|}{\multirow{2}{*}{\textbf{Model}}}  &\multicolumn{2}{|c|}{\multirow{1}{*}{\textbf{YAGO11k}}} &\multicolumn{2}{|c}{\multirow{1}{*}{\textbf{WIKIDATA12k}}} &\multicolumn{1}{|c|}{\multirow{1}{*}{\textbf{ICEWS14}}} &\multicolumn{1}{|c|}{\multirow{1}{*}{\textbf{ICEWS05-15}}} &\multicolumn{1}{|c}{\multirow{1}{*}{\textbf{GDELT100}}}\\
  &\multicolumn{1}{c}{aeIOU} &\multicolumn{1}{c|}{TAC}&\multicolumn{1}{c}{aeIOU} &\multicolumn{1}{c|}{TAC}&\multicolumn{1}{c|}{MAE}&\multicolumn{1}{c|}{MAE}&\multicolumn{1}{c}{MAE}\\
\hline
HyTE & 0.0541 & 0.0546 & 0.0541 & 0.0722 & 117.71 & 1315.46 & 122.24\\
DE-SimplE & 0.0663&0.0877&0.0484&0.0519& 83.87 & 1348.99 & 110.35 \\
TNT-Complex & 0.0840 & 0.0975 & 0.2335 & 0.2640 & 120.14 & 1281.37 & 115.97 \\
TimePlex (base) & 0.1421 & 0.1503 & 0.2620 & 0.3057 &99.58 &992.04 &109.76 \\
TimePlex & 0.2003 & 0.2253 & 0.2636 & 0.3054 &87.39 &1098.07 &102.88\\
NeuSTIP (base) & 0.1642 & - & 0.2627 & - & - & - & -\\
NeuSTIP w/ Gadgets & 0.2635 & - & 0.2630 & - & - & - & - \\
NeuSTIP w/ KGE & 0.2488 & - & 0.2735 & - & - & - & - \\
GBDT & 0.1336 & 0.1432 & 0.2923 & 0.2693 &85.81 &910.16 &94.92 \\
\hline
TEILP (event-split-td) & 0.2675 & 0.2589 & 0.3086 & 0.2995 & - & - & - \\
TEILP (event-split) & \textbf{0.2996} & 0.2861 & 0.3260 & 0.3026 &70.72 &812.07 &97.54\\
TEILP (rule-split-td) & 0.2573 & 0.2575 & 0.3228 & 0.3120 & - & - & - \\ 
TEILP (rule-split) & 0.2977 & \textbf{0.2877} & \textbf{0.3285} & \textbf{0.3153} & \textbf{70.06} &\textbf{774.01} &\textbf{94.45}\\
\hline
\end{tabular}
\caption{Time prediction performance on the benchmark datasets.}
\end{center}
\end{table*}

\subsubsection{Datasets} We evaluate the proposed method TEILP on five benchmark temporal knowledge graph datasets: WIKIDATA12k, YAGO11k \cite{dasgupta2018hyte}, ICEWS14, ICEWS05-15 \cite{garcia-duran-etal-2018-learning}, and GDELT100 \cite{leetaru2013gdelt}. According to the type of event time, we divide them into two classes: interval-based (WIKIDATA12k, YAGO11k) and timestamp-based (ICEWS14, ICEWS05-15, GDELT100). All these datasets contain temporal facts in a quadruple form, e.g., (Iran, Express intent to meet or negotiate, China, 2014-02-02). For interval-based datasets, we know both the start and end time of an event, while for timestamp-based datasets, we only know the start time. To ensure a fair comparison, we use the split provided by \cite{jain2020temporal} for WIKIDATA12k, YAGO11k, ICEWS14, ICEWS05-15 datasets and \cite{goel2020diachronic} for GDELT dataset. Note that, we delete the repeated edges in GDELT, and preserve the top 100 entities with the most edges. In the supplementary material, we provide a detailed introduction and dataset statistics. 

\subsubsection{Evaluation Metrics} For interval-based datasets, we adopt a new evaluation metric aeIOU, proposed by \cite{jain2020temporal}. It is developed from Intersection over Union (IOU), and has desirable properties for the interval time prediction task. We also use another popular metric, TAC \cite{ji2011overview} \cite{surdeanu2013overview} for evaluating intervals. The definitions are given as:
\begin{align}
   &\operatorname{aeIOU}\left(I, \widehat{I}\right) = \frac{\max \left\{1, \operatorname{vol}\left(I \cap \widehat{I}\right)\right\}}{\operatorname{vol}\left(\operatorname{ConvHull}(I, \widehat{I})\right)} \\
   &\operatorname{TAC}\left(I, \widehat{I}\right) = \frac{1}{2}\left[\frac{1}{1+|t_{s}-\widehat{t}_{s}|}+\frac{1}{1+|t_{e}-\widehat{t}_{e}|}\right]
\end{align}

\noindent where $I:=[t_s, t_e]$ denotes the ground truth, $\widehat{I}:=[\widehat{t}_s, \widehat{t}_e]$ denotes the prediction, $\text{vol}(I_a)$ represents the length of interval $I_a$, $\text{ConvHull}(I_a, I_b)$ represents the smallest single continuous interval containing both $I_a$ and $I_b$, and 1 represents the smallest time granularity. We use '1 year' for both WIKIDATA12k and YAGO11k. From (13) and (14), we know that TAC focuses on the prediction accuracy of the start and end timestamp of an interval, while aeIOU focuses on the similarity between two intervals. Both of them fall into the range of $[0,1]$, and a higher value means a better performance. For timestamp-based datasets, we follow the settings in \cite{trivedi2017know}, using Mean Absolute Error (MAE) as the evaluation metric with the smallest time granularity of '1 day'. Obviously, a lower MAE means a better model performance.

\subsubsection{Baseline Methods}
We compare TEILP\footnote{Code and data available at https://github.com/xiongsiheng/
TEILP.} with stat-of-the-art baselines for time prediction over knowledge graphs: HyTE \cite{dasgupta2018hyte}, DE-SimplE \cite{goel2020diachronic}, TNT-Complex \cite{Lacroix2020Tensor}, TimePlex \cite{jain2020temporal}, and NeuSTIP \cite{singh2023neustip}. As embedding-based methods, HyTE, DE-SimplE and TNT-Complex rank different timestamps given the known subject, object entity and relation. To obtain a time interval prediction, we adopt a greedy coalescing strategy proposed in \cite{jain2020temporal}. TimePlex is also built on embeddings, but it introduces additional temporal constraints such as relation recurrence, ordering and time gap distribution. To contrast, NeuSTIP is a temporal neuro-symbolic model which learns logical rules and Gaussian distributions from knowledge graphs. In addition, we consider gradient-boosted decision trees (GBDT), the conventional machine learning algorithm for a regression task.

\begin{table*}[ht]
\label{table-2}
\begin{center}
\begin{tabular}{l|cc|cc|c|c|c}
\hline
\multicolumn{1}{l|}{\multirow{2}{*}{\textbf{Model}}}  &\multicolumn{2}{|c|}{\multirow{1}{*}{\textbf{YAGO11k}}} &\multicolumn{2}{|c}{\multirow{1}{*}{\textbf{WIKIDATA12k}}} &\multicolumn{1}{|c|}{\multirow{1}{*}{\textbf{ICEWS14}}} &\multicolumn{1}{|c|}{\multirow{1}{*}{\textbf{ICEWS05-15}}} &\multicolumn{1}{|c}{\multirow{1}{*}{\textbf{GDELT100}}}\\
  &\multicolumn{1}{c}{aeIOU} &\multicolumn{1}{c|}{TAC}&\multicolumn{1}{c}{aeIOU} &\multicolumn{1}{c|}{TAC}&\multicolumn{1}{c|}{MAE}&\multicolumn{1}{c|}{MAE}&\multicolumn{1}{c}{MAE}\\ 
\hline
GHNN & - & - & - & - & 30.08 & 150.47 & 23.97\\
EvoKG & - & - & - & - & 29.57 & 148.17 & 23.81 \\
TimePlex (base) & 0.0700 & 0.0811 & 0.0985 & 0.1031 & 33.42 & 164.05 & 27.25 \\
TimePlex & 0.0849 & 0.0924 & 0.1120 & 0.1214 & 32.08 & 157.72 & 25.87 \\
GBDT & 0.1258 & 0.1311 & 0.1688 & 0.1771 & 26.81 & 141.38 & 22.78 \\
TEILP & \textbf{0.3175} & \textbf{0.3763} & \textbf{0.3971} & \textbf{0.4172} & \textbf{24.86} & \textbf{134.89} & \textbf{19.90} \\
\hline
\end{tabular}
\caption{Time prediction performance in future event time forecasting.}
\end{center}
\end{table*}

\subsection{Results and Analysis}

The results of the experiments are shown in Table 1, where TEILP outperforms all baselines with respect to all metrics. For our method, due to the choice of event-split or rule-split modelling, and whether to use interval duration prediction (-td), there are four versions, as noted in the table. The maximum rule length of our method is set to 5 for YAGO11k, and 3 for the others. For NeuSTIP, we use the results reported in their paper. For other baselines, we run the code on all the datasets. To deal with the incomplete events in YAGO11k and WIKIDATA12k, we remove the test queries with missing time when evaluating. 

Following conclusions can be made from the results. Conventional embedding-based methods are not suitable for time prediction since they consider it as a ranking problem similar to link prediction. Different from entities, timestamps come from a continuous space and have intrinsic connections such as order and distance. TimePlex improves its performance by adding temporal constraints. However, these constraints are still not enough for accurate time prediction. NeuSTIP introduces a similar probability distribution modelling while our method involves much more timestamps enhanced by multiple distribution types and temporally-evolving parameters. Finally, the main limitation of conventional machine learning algorithms such as GBDT is the failure to capture the complex interactions between different events.

\subsubsection{Learned Rules and Distributions} Given a query, our approach uses a chain of events, which connects subject and object entities, to reason the missing time. In particular, we focus on the events happening on either subject or object, which are of the same relation as the query or some other directly related relations. We found that the time gap between the query time and known relevant timestamps follows a certain distribution. We show some examples of learned rules and distributions here. In the YAGO11k dataset, we learn the following rule:
\begin{equation} \notag
\small
\begin{split}
& Z_{R_P, I_1, I_2, I_3}(I_q) \leftarrow E\left(F_q, F_1\right) \wedge E\left(F_1, F_2\right) \wedge E\left(F_2, F_3\right) \\
& \quad \wedge E\left(F_3, F_q\right) \wedge \text{isAffiliatedTo}\left(F_q\right)\wedge \text{isAffiliatedTo}^{-1}\left(F_1\right)  \\
 & \quad \wedge \text{isAffiliatedTo}\left(F_2\right)\wedge \text{isAffiliatedTo}^{-1}\left(F_3\right)\\
 & \quad \wedge \text{Before}(I_1, I_2) \wedge \text{Overlap}(I_2, I_3)
\end{split}
\end{equation}

Given a query $F_q = $ ('David Davis (Supreme Court justice)', 'isAffiliatedTo', 'Republican Party (United States)', $[?, ?]$), we ground the rules with $F_1 = $ ('Republican Party (United States)', 'isAffiliatedTo$^{-1}$', 'Nathaniel P. Banks', $[1857, 1875]$), $F_2 = $ ('Nathaniel P. Banks', 'isAffiliatedTo', 'Independent politician', $[1875, 1877]$), and $F_3 = $ ('Independent politician', 'isAffiliatedTo$^{-1}$', 'David Davis (Supreme Court justice)', $[1872, 1886]$). The generated conditional probability distribution is shown in Figure 3 (left) where the ground truth and our answer are $[1854, 1870]$ and $[1863, 1871]$, respectively. Similarly, given another query $F_q = $ ('John Reynolds (Canadian politician)', 'isAffiliatedTo', 'Reform Party of Canada', $[?,?]$), we ground the rules with $F_1 = $ ('Reform Party of Canada', 'isAffiliatedTo$^{-1}$', 'Raymond Speaker', $[1992, 2000]$), $F_2 = $ ('Raymond Speaker', 'isAffiliatedTo', 'Canadian Alliance', $[2000, 2003]$), and $F_3 = $ ('Canadian Alliance', 'isAffiliatedTo$^{-1}$', 'John Reynolds (Canadian politician)', $[2000, 2003]$). The generated conditional probability distribution is shown in Figure 3 (right) where the ground truth and our answer are $[1997, 2000]$ and $[1994, 2002]$, respectively. More examples are shown in the supplementary material.

\begin{figure}[htbp]
  \begin{minipage}[b]{0.23\textwidth}
    \centering
    \includegraphics[width=\textwidth]{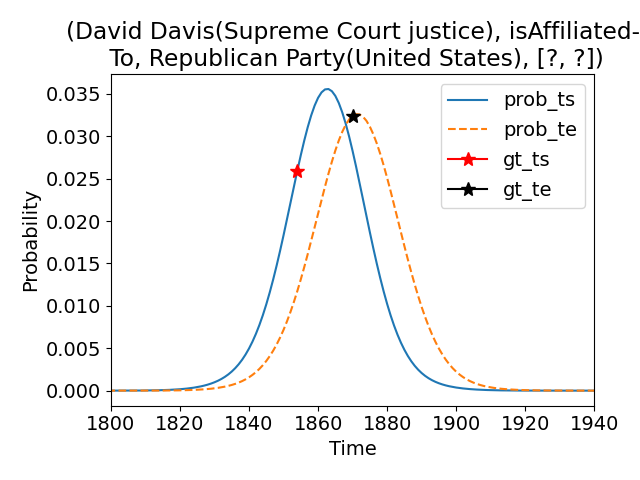}
  \end{minipage}
  \hfill
  \begin{minipage}[b]{0.23\textwidth}
    \centering
    \includegraphics[width=\textwidth]{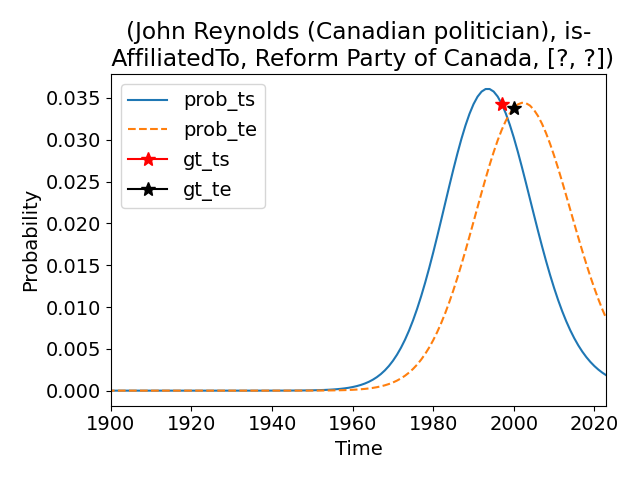}
  \end{minipage}
  \caption{The conditional probability distribution for the query interval given by TEILP.}
\end{figure}

\subsubsection{More Difficult Problem Settings} Inspired by \cite{xiong2022tilp}, we demonstrate that TEILP, which uses symbolic representations and conditional probability density functions for time prediction, is more robust than embedding-based methods. In the low-data scenario, given the same validation and test set, we change the size of the training set over a broad range. We observe a less pronounced drop in our model performance when training samples are limited. In the imbalanced scenario, given the same validation and test set, we intentionally reduce the number of training events of a certain type to investigate its effect on accuracy. We show that the learning process of TEILP is less affected by data imbalance. Further, we consider the scenario of future event forecasting which brings even more challenges to time prediction than the link prediction studied in \cite{xiong2022tilp}. In experiments, we re-split the datasets according to the start time of events. Our finding from Table 2 is that models based on temporal point process will fail with sparse training data. In contrast, both logical rules and time gap distribution modelling provide our method the generalization to unseen entities and timestamps. We provide more detailed results and analysis for all the settings in the supplementary material.

\section{Conclusion}
TEILP, an inductive logical reasoning framework, has been proposed to predict time of events in knowledge graphs. Predicting both timestamp and the interval of time can be handled by our framework. Experiments on five benchmark datasets indicate that TEILP achieves better performance than state-of-the-art methods while providing logical explanations. In addition, we consider more difficult scenarios in temporal knowledge graph reasoning, where TEILP outperforms all baseline methods. An interesting direction for future work is to predict entity attributes or event attributes which are changing with time. We need to develop efficient tools for numerical variable modelling and effectively combine them with logical rules learning, which will further extend the expressive power of neural-symbolic methods.

\section{Acknowledgments}
This work was supported by a sponsored research award by Cisco Research.

\bibliography{aaai24}

\appendix

\newpage
\section{Supplementary Material}

\subsubsection{Random Walk Sampling} For a $l$-step random walk, we start from the query, and sample next event according to the distribution described below. When the path length reaches $l$, we stop and check whether we can return to the query. We preserve the path if the returning is successful. When performing random walk, inspired by TLogic (Liu et al. 2022), we considered two transition distributions, i.e., uniform and exponential, based on the time of current event and the next one. We found that uniform distribution is better for YAGO11k and WIKIDATA12k, and exponential distribution is more suitable for ICEWS14, ICEWS05-15 and GDELT100. This is because YAGO11k and WIKIDATA12k are sparser where uniform distribution can help explore more paths. For these dense TKGs, exponential distribution sampling is focused on events that are temporally closer to the previous event and probably more relevant.

\subsubsection{Probability Density Function Design} In TEKG, the choice of probability density functions largely depends on event types. To learn them automatically, we consider three classes: i) Gaussian $\mathcal{N}(t_{q,b} - t_{i,b^{\prime}};\mu_{R_P,b,i,b^{\prime}}, \sigma_{R_P,b,i,b^{\prime}})$, ii) exponential $f_{exp}(t_{q,b} - t_{i,b^{\prime}};\lambda_{R_P,b,i,b^{\prime}})$, and iii) reflected exponential $f_{exp}(-(t_{q,b} - t_{i,b^{\prime}});\lambda_{R_P,b,i,b^{\prime}})$, where subscript $b=\{\text{'s'}, \text{'e'}\}$, index $i \in \{1, \cdots, l\}$, and subscript $b^{\prime}=\{\text{'s'}, \text{'e'}\}$. Further, $\mathcal{N}(\cdot)$ denotes the probability density function of a Gaussian distribution with parameter $\mu$ and $\sigma$, and $f_{exp}(\cdot)$ denotes the probability density function of an exponential distribution with parameter $\lambda$. The one that best fits the training samples will be selected.

Further, we consider the distribution parameters $\mu$, $\sigma$ and $\lambda$ conditioned on rule pattern $R_P$, subscript $b$, index $i$ and subscript $b^{\prime}$. To be specific, $R_P$, $b$, $i$ and $b^{\prime}$ are all considered as categorical variables. We also consider these parameters evolving with time. To be specific, given $R_P$, $b$, $i$ and $b^{\prime}$, the distribution parameters $\mu$, $\sigma$ and $\lambda$ are estimated separately in different time ranges. In WIKIDATA12k and YAGO11k, we split the entire time range into equal-length pieces. Each piece has a length of '100 years', e.g., $[1800, 1900]$ and $[1900, 2000]$. In other datasets, we do not split since the entire time ranges are relatively short. We estimate these parameters during training. Given the definition of the probability density functions $f_{R_P, b, i, b^{\prime}}(\cdot)$, we obtain $g_{R_P,b,i}()$ and $G_{R_P,b}()$ with (3) and (4). Figure 4 shows an example of the conditional probability distribution given by TEILP.


\begin{table*}[!ht]
\label{table-3}
\begin{center}
\begin{tabular}{l|ccccccc}
\hline
Dataset  & $|\mathcal{G}_{train}|$ & $|\mathcal{G}_{valid}|$ & $|\mathcal{G}_{test}|$ & $|\mathcal{E}|$ & $|\mathcal{R}|$ & $|\mathcal{T}|$ & $|\mathcal{I}|$
\\ 
\hline
WIKIDATA12k & 32,497 & 4,062 & 4,062 & 12,544 & 24 & 237 & 2,564\\
YAGO11k & 16,408 & 2,051 & 2,050 & 10,622 & 10 & 251 & 6,651\\
ICEWS14 & 72,826 & 8,941 & 8,963 & 7,128 & 230 & 365 & - \\
ICEWS05-15 & 386,962 & 46,275 & 46,092 &  10,488 & 251 & 4017 & -\\
GDELT100 & 390,045 & 48,756 & 48,756 & 100 & 20 & 365 & -\\
\hline
\end{tabular}
\caption{Dataset statistics for the five benchmark datasets}
\end{center}
\end{table*}

We compare our approach with the existing methods developed for time prediction in TKGs: Know-Evolve \cite{trivedi2017know}, TimePlex \cite{jain2020temporal}, and TELLER \cite{li2021explaining}. Both Know-Evolve and TELLER adopt a temporal point process which predicts the time of the next event given all the past events. Their limitation in the general setting is that given a query, we can hardly decide which events in the TKG should be used as past events. In addition, those events in the more distant future, which are also helpful in time prediction over TKGs, are ignored in these methods. On the other hand, TimePlex introduces the embedding of timestamps as well as additional soft temporal constraints for time prediction. Embedding-based methods will fail for unseen timestamps. Further, these soft temporal constraints only use a single Gaussian distribution for relation-pairwise time gap modelling. Different distribution types, various relational patterns and the evolving of distribution parameters are ignored in TimePlex. 

\begin{figure}[t]
    \centering
    \includegraphics[width=0.47\textwidth]{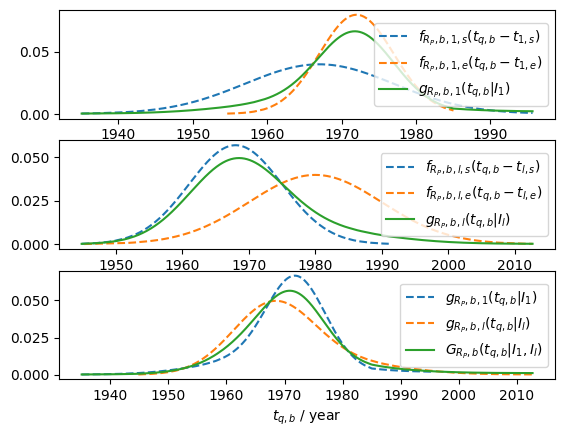}
  \caption{The conditional probability distribution given by TEILP.}
\end{figure}

\subsubsection{Rule Learning} We introduce more details for conditional probability matrix calculation and the LSTM model setting.
Conditional probability matrix $\mathbf{c}_{q,b,L}\in[0,1]^{|\mathcal{F}|\times|S_t|}$ is based on the conditional probability density function $g_{R_P, b, l}(\cdot)$, where $\mathcal{F}$ denotes the set of facts, and $S_t$ denotes the set of quantized timestamps. We define its $(m,r)$ entry $(\mathbf{c}_{q,b,L})_{m,r}$ as $\frac{1}{|S^{\prime}_{R_P}|}\sum^{|S^{\prime}_{R_P}|}_{\kappa = 1}{g}_{R^{\kappa}_P,b,l_{\kappa}}(t_r|I_m)$, where $t_r$ denotes a candidate timestamp in $S_t$, and $I_m$ denotes the interval of event $F_m$. Further, $S^{\prime}_{R_P}$ denotes the set of rule patterns that $F_q$ and $F_m$ satisfy, i.e., $\exists I_1, \cdots, I_{l_{\kappa}-1}$, s.t. $Z_{R^{\kappa}_P, I_1, \cdots, I_{l_{\kappa}-1}, I_m}(I_q) = 1$. Given the rule pattern $R^{\kappa}_P$, we require $F_m$ to be the last event on the path, where $l_\kappa$ denotes the path length. Essentially, $\mathbf{u}_{L+1}$ denotes the probability of different events after at most $L$-step random walk that we find as the last event, and given the intervals of the events, $\mathbf{c}_{q,b,L}$ shows the conditional probability of the query timestamp $t_{q,b}$.

Based on previous analysis, we know that the task of learning temporal logical rules for time prediction is equivalent to learning attention vectors $\mathbf{\alpha, \beta, \gamma}$ which softly select predicates, temporal relations and rule lengths, respectively. Instead of optimizing these parameters separately, we consider building the connection between different steps. Inspired by TILP \cite{xiong2022tilp}, we use an LSTM model to ensure that current step's attention vectors depend on previous steps', i.e.,

\begin{align}
\mathbf{h}_{i} &= \operatorname{update} \left(\mathbf{h}_{i-1}, \mathbf{x}_i\right) \\
\mathbf{\alpha}^{i} &= \operatorname{softmax}\left(W_{TR} \mathbf{h}_{i}+b_{TR}\right)\\
\mathbf{\beta}^{i} &= \operatorname{softmax}\left(W_P \mathbf{h}_{i}+b_P\right)
\end{align}

\begin{align}
\mathbf{\gamma}^{i} &=\operatorname{softmax}\left(\left[\mathbf{h}_{0}, \ldots, \mathbf{h}_{i-1}\right]^T \mathbf{h}_{i}\right)
\end{align}

\noindent where $\mathbf{h}_{i}\in \mathbb{R}^{d}$ denotes the $i$th-step hidden state with dimension $d$, and $\mathbf{h}_{0}$ is set as $\mathbf{0}$. Further, $\mathbf{x}_{i}\in \mathbb{R}^{d^{\prime}}$ denotes the $i$th-step input with dimension $d^{\prime}$. For $1\leq i\leq L$, $\mathbf{x}_{i}$ is the embedding of the query predicate $P_q$, and $\mathbf{x}_{L+1}$ is the embedding of an 'END' token. Finally, $W_P\in\mathbb{R}^{|P|\times d}$, $W_{TR}\in\mathbb{R}^{|TR|\times d}$, $b_P\in\mathbb{R}^{|P|}$, $b_{TR}\in\mathbb{R}^{|TR|}$ are all learnable parameters. Figure 2 illustrates the rule learning system.

\subsubsection{Acceleration}
Efficient learning on large TKGs is a realistic task especially for complex models. In this paper, we use several strategies to accelerate our algorithm. Before the introduction of these strategies, we first analyze the factors that affect the time complexity of our algorithm. The first influencing factor is the size of the graph. In TEKG, the number of event nodes decides the dimension of matrix operators $M_P$ and $M_{E,TR}$. The graph size increases significantly for large TKGs. To solve this problem, we use a \textbf{local} graph instead of the global one for each batch of samples. To be specific, given a batch of samples, we first perform random walk to obtain their neighbors. Then all the following computations happen on this local graph. Another influencing factor is the number of positive examples used for training. Our solution is to randomly select a fixed number of training samples for each target predicate. Finally, we simultaneously learn the structure and confidence of all the logical rules with the framework. The number of rules used in application is also an influencing factor. We ignore those low-confidence rules, and apply only the selected rules to obtain the local graph for test queries.

\subsubsection{Efficiency Study} The worst-case time complexity for the conversion of TEKG is $\mathcal{O}(|\mathcal{F}|^2 + |\mathcal{T}|)$, where $\mathcal{F}$ denotes the set of event nodes, and $\mathcal{T}$ denotes the set of timestamp nodes. To extract the rule pattern candidates and fit the conditional probability density functions, the time complexity is $\mathcal{O}(|\mathcal{P}|N_{pw}LD_{max})$ in the worst case, where $\mathcal{P}$ is the set of predicates, $N_{pw}$ is the number of sampling walks for a single predicate, $L$ is the maximum rule length, and $D_{max}$ is the maximum event node degree in TEKG. Similarly, given a query, the time complexity for path sampling to obtain a local graph is $\mathcal{O}(N_{qw}LD_{max})$ in the worst case, where $N_{qw}$ is the number of walks for a single query. Given a query event $F_q$, the time complexity for probability distribution calculation is $\mathcal{O}(|S_t|N_{R_P, P_q}N_{path, R_P})$ in the worst case, where $S_t$ is the set of quantized timestamps, $N_{R_P, P_q}$ is the number of rules given the query predicate $P_q$, and $N_{path,R_P}$ is the max number of paths given rule pattern $R_P$. Finally, the worst-case time complexity for differentiable temporal random walk is $\mathcal{O}(|\mathcal{F}^{\prime}|^2L + |\mathcal{F}^{\prime}|^2|S_t||)$, where $\mathcal{F}^{\prime}$ is the set of event nodes in the local graph.

\subsubsection{Dataset Statistics} WIKIDATA12k is a subgraph extracted from WIKIDATA dataset with temporal information \cite{dasgupta2018hyte}. YAGO11k, formed from YAGO3 dataset \cite{dasgupta2018hyte}, is another large temporal knowledge graph with time intervals. ICEWS14 and ICEWS05-15 are two event-based temporal knowledge graphs with different time ranges from the Integrated Crisis Early Warning System \cite{garcia2018learning}. GDELT is also an event-based temporal knowledge graph extracted from Global Database of Events, Language, and Tone \cite{trivedi2017know}. The relations in WIKIDATA12k and YAGO11k, e.g., 'member of', 'residence', 'worksAt', are time-sensitive. The temporal information is also important for the relations in ICEWS14, ICEWS05-15, and GDELT, e.g., 'Make statement', 'Protest violently, riot', 'Occupy territory'. Table 3 shows the statistics of the datasets used in this paper, where $\mathcal{G}, \mathcal{E}, \mathcal{R}, \mathcal{T}, \mathcal{I}$ denote the set of edges, entities, relations, timestamps and intervals, respectively.

\subsubsection{Learned Rules and Distributions} We show more examples of learned rules and distributions here. In the YAGO11k dataset, we learn the following rule:
\begin{equation} \notag
\begin{split}
Z_{R_P, I_1, I_2}(I_q) & \leftarrow E\left(F_q, F_1\right) \wedge E\left(F_1, F_2\right) \wedge E\left(F_2, F_q\right)\\ 
& \wedge \text{wasBornIn}\left(F_q\right) \wedge \text{wasBornIn}^{-1}\left(F_1\right) \\
&\wedge \text{isMarriedTo}^{-1}\left(F_2\right) \wedge \text{Before}(I_1, I_2)
\end{split}
\end{equation}

Given a query $F_q = $ ('Jiang Qinqin', 'wasBornIn', 'China', $[?, ?]$), we ground the rules with $F_1 = $ ('China', 'wasBornIn$^{-1}$', 'Chen Jianbin', $[1970, 1970]$), and $F_2 = $ ('Chen Jianbin', 'isMarriedTo$^{-1}$', 'Jiang Qinqin', $[2006, 2006]$). The generated conditional probability distribution is shown in Figure 5 (left) where the ground truth and our answer are $[1975, 1975]$ and $[1976, 1976]$, respectively. Similarly, given another query $F_q = $ ('Asif Ali Zardari', 'wasBornIn', 'Karachi', $[?,?]$), we ground the rules with $F_1 = $ ('Karachi', 'wasBornIn$^{-1}$', 'Benazir Bhutto', $[1953, 1953]$), and $F_2 = $ ('Benazir Bhutto', 'isMarriedTo$^{-1}$', 'Asif Ali Zardari', $[1987, 2007]$). The generated conditional probability distribution is shown in Figure 5 (right) where the ground truth and our answer are $[1955, 1955]$ and $[1957, 1957]$, respectively..

\begin{figure}[htbp]
  \begin{minipage}[b]{0.23\textwidth}
    \centering
    \includegraphics[width=\textwidth]{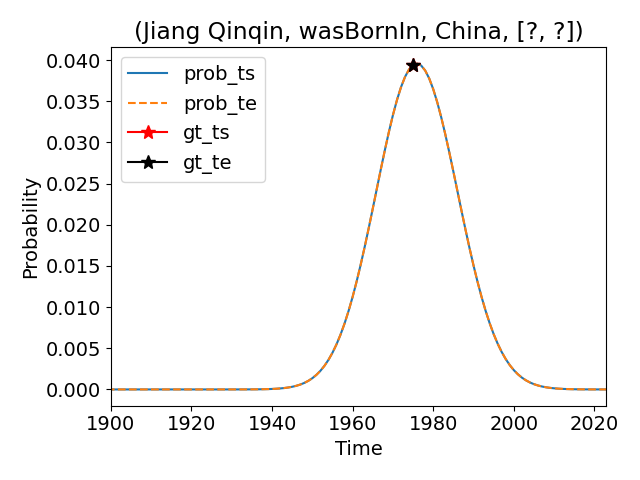}
  \end{minipage}
  \hfill
  \begin{minipage}[b]{0.23\textwidth}
    \centering
    \includegraphics[width=\textwidth]{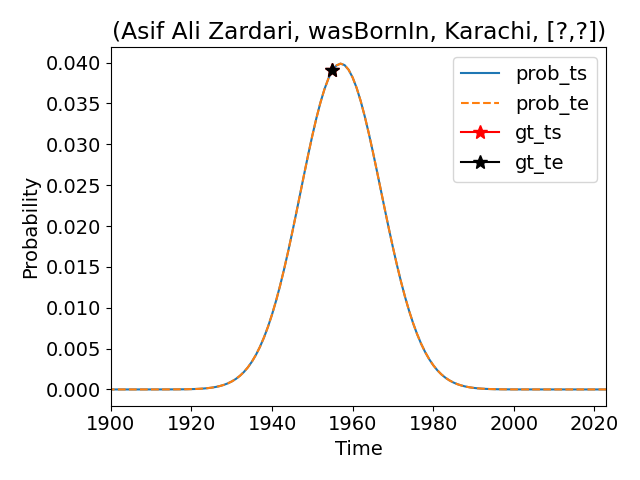}
  \end{minipage}
  \caption{The conditional probability distribution for the query interval given by TEILP.}
\end{figure}

Likewise, we learn the following rule in the WIKIDATA12k dataset:
\begin{equation} \notag
\begin{split}
& Z_{R_P, I_1, I_2, I_3}(I_q) \leftarrow E\left(F_q, F_1\right) \wedge E\left(F_1, F_2\right) \wedge E\left(F_2, F_3\right) \\ &\quad \quad  \wedge E\left(F_3, F_q\right) \wedge \text{SignificantEvent}\left(F_q\right)  \\
&\quad \quad \wedge \text{SignificantEvent}^{-1}\left(F_1\right) \wedge \text{SignificantEvent}\left(F_2\right)\\
&\quad \quad\wedge \text{SignificantEvent}^{-1}\left(F_3\right) \wedge \text{Overlap}(I_1, I_2) \\
&\quad \quad \wedge \text{Before}(I_2, I_3)
\end{split}
\end{equation}

Given a query $F_q = $ ('Empire State Building', 'significant event', 'construction', $[?, ?]$), we ground the rules with $F_1 = $ ('construction', 'significant event$^{-1}$', 'Woolworth Building', $[1910, 1913]$), $F_2 = $ ('Woolworth Building', 'significant event', 'opening', $[1913, 1913]$), and $F_3 = $ ('opening', 'significant event$^{-1}$', 'Empire State Building', $[1931, 1931]$). The generated conditional probability distribution is shown in Figure 6 (left) where the ground truth and our answer are $[1930, 1931]$ and $[1930, 1931]$, respectively. Similarly, given another query $F_q = $ ('USS Leahy', 'significant event', 'keel laying', $[?,?]$), we ground the rules with $F_1 = $ ('keel laying', 'significant event$^{-1}$', 'USS Sennet', $[1944, 1944]$), $F_2 = $ ('USS Sennet', 'significant event', 'ship commissioning', $[1944, 1944]$), and $F_3 = $ ('ship commissioning', 'significant event$^{-1}$', 'USS Leahy', $[1968, 1968]$). The generated conditional probability distribution is shown in Figure 6 (right) where the ground truth and our answer are $[1959, 1959]$ and $[1964, 1965]$, respectively.

\begin{figure}[tbp]
  \begin{minipage}[b]{0.23\textwidth}
    \centering
    \includegraphics[width=\textwidth]{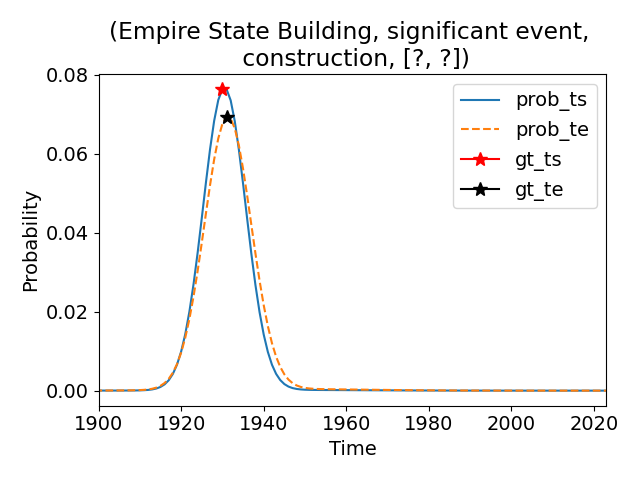}
  \end{minipage}
  \hfill
  \begin{minipage}[b]{0.23\textwidth}
    \centering
    \includegraphics[width=\textwidth]{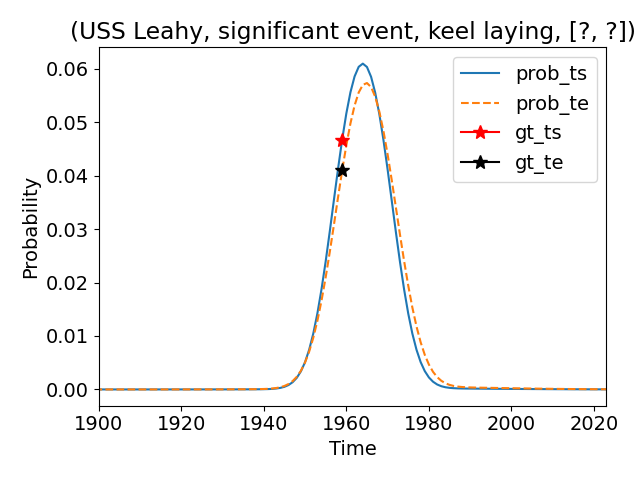}
  \end{minipage}
  \caption{The conditional probability distribution for the query interval given by TEILP.}
\end{figure}

\subsubsection{More Difficult Problem Settings} Compared with the original one, these settings make the problem more difficult and thus hurt the model performance. We proved that our method benefited from inductive logical reasoning is more robust than baselines. When considering a specific setting, we should investigate the dataset and analyze the performance. For example, we can slightly reduce the training set size to check data sufficiency. Similarly, separately reducing the training samples of different relations is useful for imbalance issue test. Moreover, time range estimation of test query can help decide whether we are predicting future events based on past facts.

\subsubsection{Few Training Samples} In applications such as personalized medicine and rare diseases research, training data are limited and costly to obtain. The model performance with restricted training samples becomes vital. We conduct a parametric performance analysis of different models in such a low-data scenario. In our experiments, we randomly select samples within the original training set and evaluate various models on WIKIDATA12k and YAGO11k datasets. To obtain reliable results, given a fixed ratio of the training set size, we repeat the experiment for 5 rounds. We show the average aeIOU and TAC curves with err bars in Figure 7. We observe that TEILP outperforms all the baselines when the training set size decreases. Different from embedding-based methods, the inductive logical reasoning approach TEILP learns rules and distributions which are independent on entities and events. A small training set only affects the frequency of different rule patterns and the distribution parameter estimation. This helped TEILP to outperform embedding-based methods. In contrast, TimePlex needs a large amount of data in the embedding learning of various entities, relations and timestampes. Regression-based method GBDT also fails on YAGO11k where there exist complex interactions between events.

\begin{figure}[!bp]
  \begin{minipage}[b]{0.23\textwidth}
    \centering
    \includegraphics[width=\textwidth]{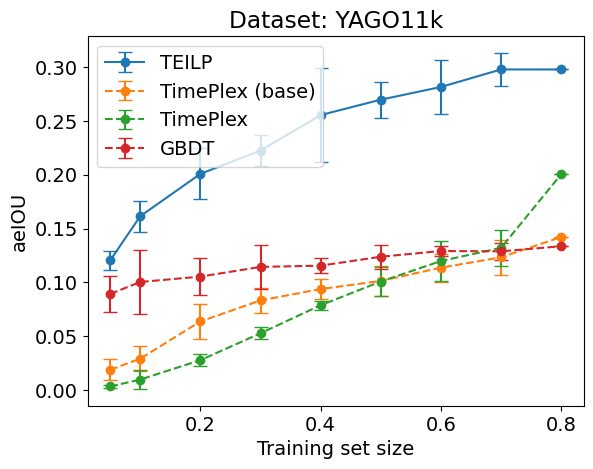}
  \end{minipage}
  \hfill
  \begin{minipage}[b]{0.23\textwidth}
    \centering
    \includegraphics[width=\textwidth]{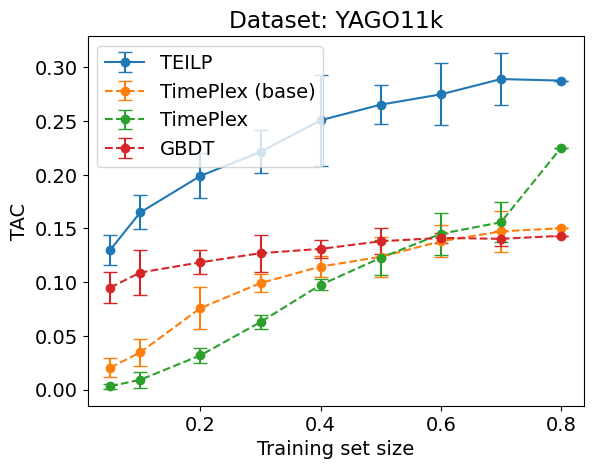}
  \end{minipage}
    \begin{minipage}[b]{0.23\textwidth}
    \centering
    \includegraphics[width=\textwidth]{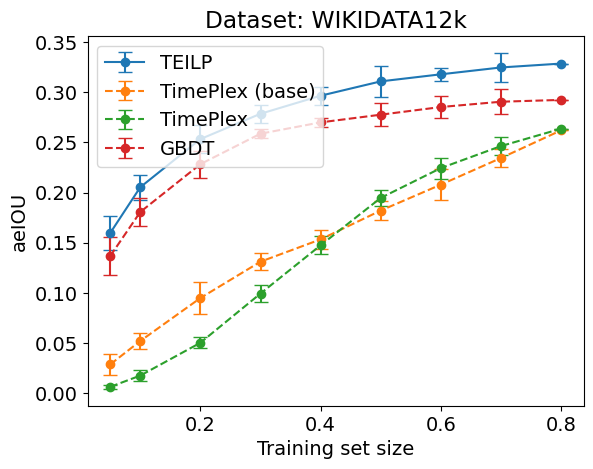}
  \end{minipage}
  \hfill
  \begin{minipage}[b]{0.23\textwidth}
    \centering
    \includegraphics[width=\textwidth]{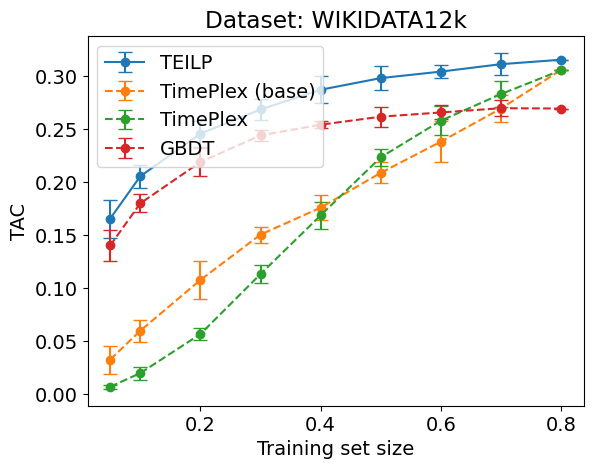}
  \end{minipage}
  \caption{Time prediction performance with the few training samples setting.}
\end{figure}

\subsubsection{Imbalanced Event Types} Event type imbalance is a common problem in large TKGs. For example, there are 40621 events of 24 types in WIKIDATA12k. The most frequent type 'member of sports team' has 16090 samples, while the number of events for relation 'capital of' and 'residence' are only 86 and 80, respectively. In experiments, we first construct a balanced test set to ensure a fair evaluation. Given the entire dataset, the ratio for test set is $10\%$. In the ideal balanced test set, each event type has the same number of samples, i.e., $1\%$ for YAGO11k and $0.417\%$ for WIKIDATA12k. For rare event types which have less samples than the ideal ratio, we randomly choose half of them for test. Then we construct the imbalanced training set by randomly reducing $50\%$ of the events for a specific type. We calculate the performance change affected by this operation. To obtain reliable results, given a specific imbalanced event type, we repeat the experiment for 5 rounds. We show the average aeIOU and TAC changes in Figure 8. For example, for relation id 4 'isMarriedTo' in YAGO11k, the average aeIOU change of TEILP is $65.0\%$. It means the aeIOU for relation 'isMarriedTo' after reducing $50\%$ of the same type training events is $65.0\%$ of that before the reducing operation. To contrast, the average aeIOU change of TimePlex (base), TimePlex and GBDT for the same relation are $35.5\%$, $12.3\%$ and $64.2\%$, respectively. We found that in TEILP the attention vectors of predicates, temporal relations and rule length are dependent on the event type, making it less affected by data imbalance. In contrast, embeddings of entities are shared among all the relations, making embedding-based method more susceptible to event type imbalance.

\begin{figure}[!tbp]
  \begin{minipage}[b]{0.23\textwidth}
    \centering
    \includegraphics[width=\textwidth]{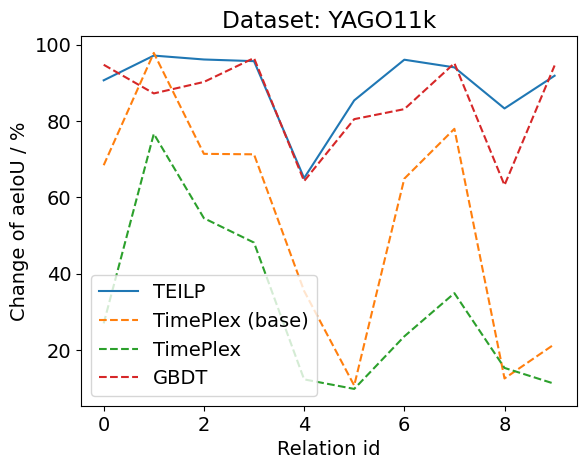}
  \end{minipage}
  \hfill
  \begin{minipage}[b]{0.23\textwidth}
    \centering
    \includegraphics[width=\textwidth]{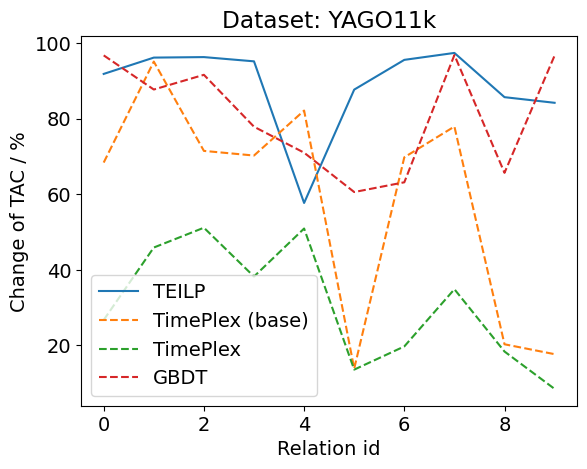}
  \end{minipage}
    \begin{minipage}[b]{0.23\textwidth}
    \centering
    \includegraphics[width=\textwidth]{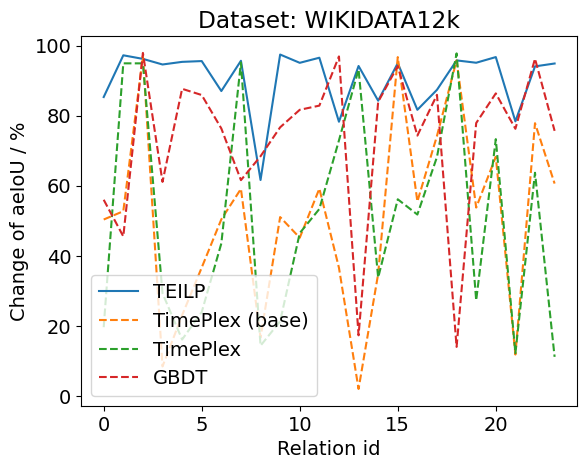}
  \end{minipage}
  \hfill
  \begin{minipage}[b]{0.23\textwidth}
    \centering
    \includegraphics[width=\textwidth]{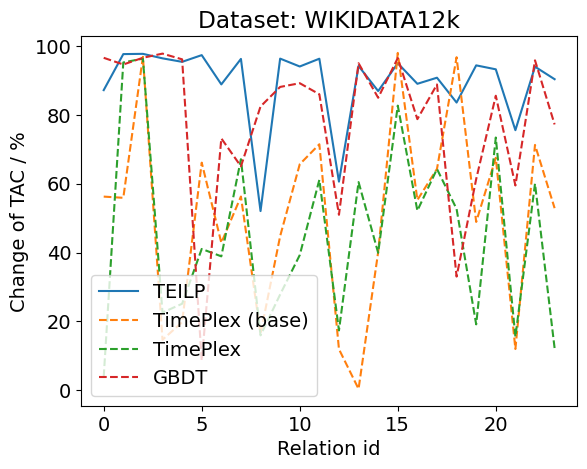}
  \end{minipage}
  \caption{Time prediction performance with the imbalanced event types setting.}
\end{figure}

\subsubsection{Future Event Time Forecasting} A typical scenario for the time shifting problem in TKGs is to forecast the time of future event based on past events. In experiments, we first re-split the datasets according to the start time of events. To be specific, the start time range for training, validation and test in WIKIDATA12k are $[-3, 2009]$, $[2009, 2012]$, $[2012, 2018]$, respectively. The start time range for training, validation and test in YAGO11k are $[-431, 2007]$, $[2007, 2011]$, $[2011, 2023]$, respectively. Note that, both WIKIDATA12k and YAGO11k use a resolution of '1 year'. Similarly, the time range for training, validation and test in ICEWS14 are ['2014/01/01', '2014/10/24'], ['2014/10/24', '2014/11/24'], ['2014/11/24', '2014/12/31'], respectively. The time range for training, validation and test in ICEWS05-15 are ['2005/01/01', '2013/11/18'], ['2013/11/18', '2014/12/28'], ['2014/12/28', '2015/12/31'], respectively. The time range for training, validation and test in GDELT100 are ['2015/04/01', '2016/01/26'], ['2016/01/26', '2016/02/28'], ['2016/02/28', '2016/03/31'], respectively. Note that, ICEWS14, ICEWS05-15 and GDELT100 all use a resolution of '1 day'.

The task is to predict the time gap between last relevant event and next event. Depending on the type of relations and sparsity of data, we define the relevant events as the one happening on the same subject / object. Note that relevant events in sparse TKGs such as WIKIDATA12k and YAGO11k are defined as any relation types, while in dense TKGs such as ICEWS14, ICEWS05-15 and GDELT100 are defined as the same relation types. We found that for sparse TKGs, events are usually low-frequent and last for years, while in dense TKGs, events happen every few days.

To deal with the time range variance, given a query, we only use past events to learn logical rules and probability distributions of the time gap. This strategy is also applied to other baselines (GHNN \cite{han2020graph}, EvoKG \cite{park2022evokg}, TimePlex and GBDT) to obtain a fair comparison. Both GHNN and EvoKG model TKGs as a temporal point process with dynamic embeddings. Due to their application constraints, we only evaluate them on timestamp-based datasets. Note that, we adopt a more \textbf{strict} setting where previous events in the validation and test set will not be used for forecasting. The results are shown in Table 2 where our method is much better than the baselines. We observe that models based on temporal point process will fail if not all the previous events are provided. In contrast, both logical rules and time gap distribution modelling provide our method the generalization to unseen timestamps.

\end{document}